\documentclass[letterpaper, 10 pt, conference]{ieeeconf}  %

\IEEEoverridecommandlockouts                              %

\usepackage{graphics} %
\usepackage{times} %
\usepackage{amsmath} %
\usepackage{amssymb}  %
\usepackage{amsfonts}
\usepackage{subfigure}
\usepackage{hyperref}
\usepackage{graphicx}
\usepackage{float}
\usepackage{caption}
\usepackage{multirow}
\usepackage{color}

\title{\LARGE \bf
TacIPC: Intersection- and Inversion-free FEM-based Elastomer Simulation For Optical Tactile Sensors
}

\author{Wenxin Du$^{1*}$, Wenqiang Xu$^{1*}$, Jieji Ren$^{1}$, Zhenjun Yu$^{1}$ and Cewu Lu$^{1}$%
\thanks{*Equal contribution.}%
\thanks{$^{1}${\tt\small \{mnkmYuki, vinjohn, jiejiren, jeffson-yu, lucewu\}@sjtu.edu.cn}. Cewu Lu is the corresponding author, a member of Qing Yuan Research Institute and MoE Key Lab of Artificial Intelligence, AI Institute, Shanghai Jiao Tong University, China.}%
}

\begin{document}

\maketitle
\thispagestyle{empty}
\pagestyle{empty}

\begin{abstract}

Tactile perception stands as a critical sensory modality for human interaction with the environment. Among various tactile sensor techniques, optical sensor-based approaches have gained traction, notably for producing high-resolution tactile images. This work explores gel elastomer deformation simulation through a physics-based approach. While previous works in this direction usually adopt the explicit material point method (MPM), which has certain limitations in force simulation and rendering, we adopt the finite element method (FEM) and address the challenges in penetration and mesh distortion with incremental potential contact (IPC) method. As a result, we present a simulator named TacIPC, which can ensure numerically stable simulations while accommodating direct rendering and friction modeling. To evaluate TacIPC, we conduct three tasks: pseudo-image quality assessment, deformed geometry estimation, and marker displacement prediction. These tasks show its superior efficacy in reducing the sim-to-real gap. Our method can also seamlessly integrate with existing simulators. More experiments and videos can be found in the supplementary materials and on the website: \url{https://sites.google.com/view/tac-ipc}.

\end{abstract}

\section{Introduction}
\begin{figure}[t!]
    \centering
    \includegraphics[width=1\linewidth]{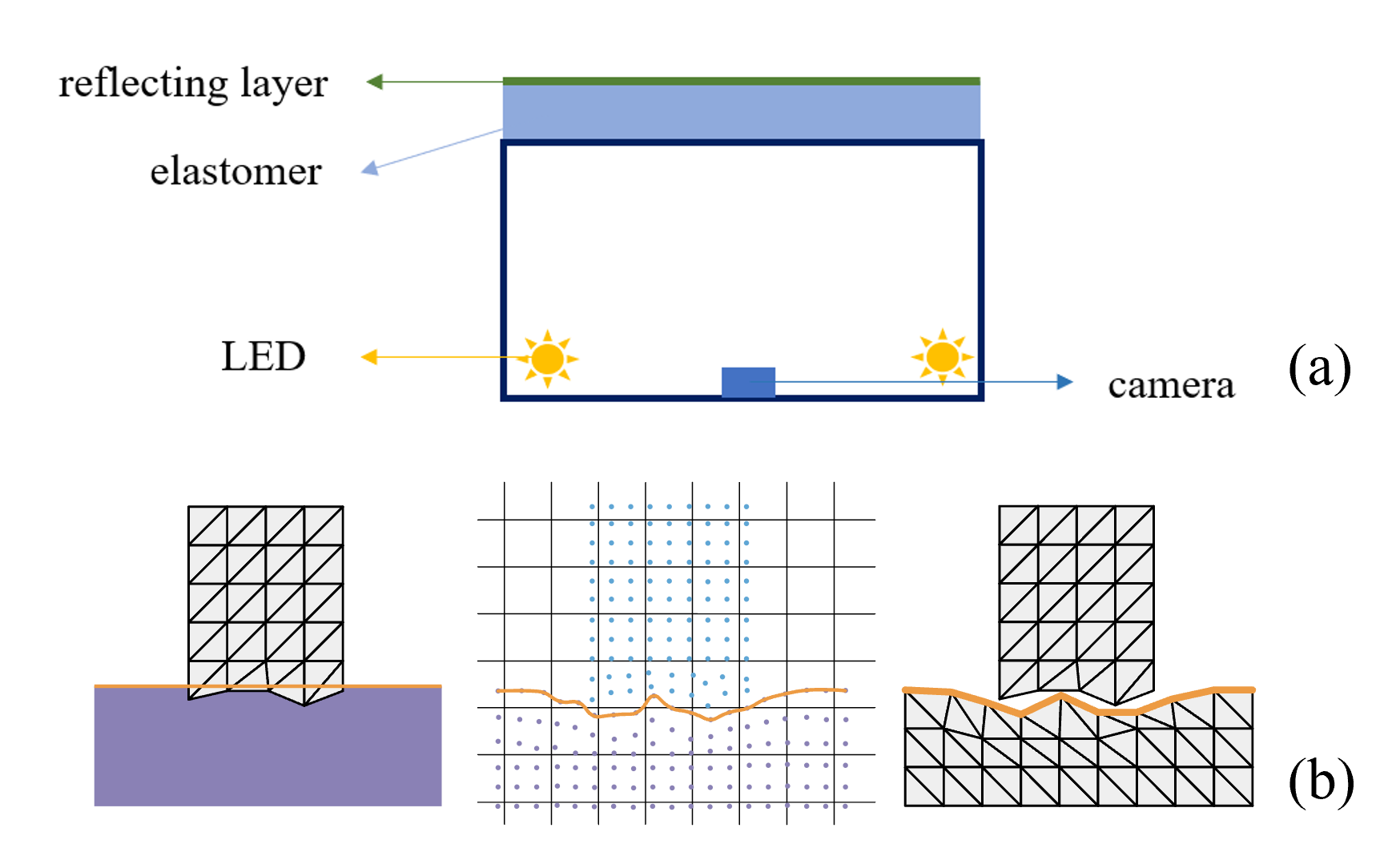}
    \caption{(a). The structure of optical tactile sensors. (b). Different ways to implement elastomer: non-physics, MPM-based, FEM-based.}
    \label{fig:fig1}
\end{figure}

Tactile perception plays a pivotal role in enabling humans to discern and interact with their surrounding environment. Several methods have been explored to implement tactile sensors, including piezoelectric sensors \cite{piezoelectric}, capacitive sensors \cite{capacitive}, and optical sensors \cite{mctac,gelsight,gelslim}. Recently, optical sensor-based approaches have become more and more popular due to their ability to generate high-resolution tactile images (also known as pseudo images) and their seamless integration into learning pipelines. The optical tactile sensors often use a camera to capture the deformation of the gel elastomer (as shown in Fig. \ref{fig:fig1}). However, the tactile images derived from optical sensors do not directly correspond to force measurements. Many carefully designed light paths are proposed in previous sensors \cite{mctac,gelsight,gelslim}, but none have been able to accurately and consistently reconstruct the deformed surface normals across various shapes. Consequently, the primary application of the optical tactile sensor is its direct incorporation into a learning pipeline \cite{vtaco}, with simulation being an integral component of this process. 

For simulation, the simulator should be able to support the computation and rendering of the gel elastomer deformation. In the previous works, the gel deformation is treated in two directions: \textit{non-physics-based} and \textit{physics-based} ways. As shown in Fig. \ref{fig:fig1}, the non-physical way \cite{depth1,depth2} allows the object and the gel to penetrate each other, and uses the penetration depth as the elastomer deformation. 
The physics-based way \cite{tacchi,mpm2} usually adopts the explicit material point method (MPM) as the physics simulation algorithm. The object and gel are both modeled as sets of material points. Explicit MPM benefits from the development of Taichi \cite{taichi}, which has friendly APIs, thus simple coding is enough to build effective simulators. However, explicit MPM has its disadvantages over force simulation (especially frictional force) and its stability (particle penetration and numeric instability), and the rendering is implemented by interpolation between sample points, which can lead to over-smooth effects. Another way to achieve physics-based deformation is the classic finite element method (FEM)-based modeling, which is known to easily model friction and accurate force computation. However, since FEM adopts explicit mesh for the deformed object (i.e., the elastomer), the contact between the object and the gel can easily cause penetration and mesh distortion \cite{distortion}.

In this work, we take the FEM-based path to model the elastomer. The penetration and mesh distortion issues are addressed by incorporating the incremental potential contact (IPC) method \cite{ipc}, which can guarantee intersection-free and inversion-free during contact. In addition, IPC solves the dynamics equation in an implicit way, leading to numerically stable and accurate simulation results. The proposed simulator is named \textbf{TacIPC}. It models the elastomer with a tetrahedral mesh. Explicit mesh modeling has three advantages: (1) Rendering can be directly handled. (2) Friction can be directly modeled. (3). It can be naturally integrated with mesh-based robot simulators such as MuJoCo \cite{mujoco}, Bullet \cite{pybullet}, Isaac Sim \cite{isaacgym}, and RFUniverse \cite{rfuniverse}. 

For rendering, in the real world, the image is generated by the reflecting layer. The typical Phong-based rendering with Gaussian blur \cite{depth1,tacto} does not replicate the mechanism and only uses it to create seemingly similar effects. In TacIPC, we adopt physics-based ray-tracing rendering to mimic the light path as in the real world by adjusting the LED positions. We calibrate the lighting color temperature in the real world and implement it in the simulation. Also, we use reference images from the real world to generate real-world sensory noise.

To evaluate the proposed tactile simulator, we conduct three tasks: pseudo-image quality assessment, marker displacement prediction, and deformed geometry estimation. These tasks can demonstrate that compared with previous optical tactile sensor simulators, our simulation method can significantly reduce the sim-to-real gap in both rendering and physics simulation, providing a more accurate and reliable way to design and optimize gel-based tactile sensors or provide training data for learning methods. 
Furthermore, our approach can be easily integrated into existing simulation environments and can significantly improve the performance and accuracy of the simulation. We adopt RFUniverse \cite{rfuniverse} in our experiments.

We summarize our contribution as follows:
\begin{itemize}
    \item We propose the first intersection- and inversion-free simulator TacIPC for optical tactile sensors. It utilizes the physical properties of the sensor to simulate realistic tactile images in a numerically stable and physically accurate way. 
    \item We use TacIPC to predict the tactile sensor marker displacement under frictional contacts, which is essential for contact-rich manipulation tasks. It significantly outperforms the MPM-based simulator. 
    \item We train a depth estimation neural network in a sim-to-real fashion using pseudo images generated by TacIPC and validate the model on real-world data.
\end{itemize}

\section{Related Works}
Our work proposes a physics-based simulator for optical tactile sensors which is related to techniques of simulating the elastomer deformation and the tactile sensors.

\subsection{Elastomer Simulation}
Previous methods to simulate the elastomer effect in optical tactile sensors can be roughly divided into two categories: \textit{non-physics-based} and \textit{physics-based} approaches.

The non-physics-based approaches try to produce realistic tactile images in the simulator without actually considering the elastomer deformation caused by external contact. Some works \cite{depth1,depth2} reconstruct the pseudo elastomer surfaces by allowing the object to intersect the elastomer and take the depth of the intersected surface. Then, they apply Gaussian kernels to smooth depth maps. We refer to these methods as ``depth-based''. TACTO \cite{tacto} is also one of the depth-based methods to render the tactile image with PyBullet \cite{pybullet}. But it just adds shadow on the intersected surface. These non-physics-based approaches can produce seemingly plausible pseudo images but they do not have physical context.

Another line is the physics-based approach. Simulating elastomers, such as the gel layers in optical tactile sensors, is a complex task due to their unique deformable property and high degrees of freedom. Numerous studies have adopted the Finite Element Method (FEM) for deformation simulations, as illustrated in \cite{fem1,zhang2022hardware}. In \cite{zaidi2017model}, the authors proposed a robot interface to grasp an elastomer such as rubber spheres and foam cubes. Lee et al. \cite{fem1} use FEM to estimate force distribution. Though FEM is widely adopted in simulating deformation, it is considered computational consumption and likely causes mesh distortion (e.g., irregular mesh structures or negative element volumes) \cite{distortion}. Recently, several works have leveraged the material point method (MPM) for a physics-based simulation of elastomers \cite{mpm2,tacchi}. Such methods enjoy the friendly API provided by Taichi \cite{taichi}, and can easily utilize the explicit MPM method to develop simulators. However, explicit MPM has its disadvantages over frictional force simulation and numeric stability. And the particle-based representation makes it hard to render the tactile image directly. Thus, by rethinking the trend of technical development, we determine physical accuracy is more important and take the FEM path. In this work, we address the distortion issue of FEM simulation by introducing increment potential contact (IPC) \cite{ipc}, which can guarantee intersection- and inversion-free.

\subsection{Sensor Simulation}
With the elastomer simulation technique, we need an interface to incorporate it into a sensor and robot system. For example, in the non-physics-based line of research, \cite{depth1} initially simulated optical tactile sensors by refining depth maps sourced from Gazebo \cite{gazebo}. Meanwhile, in \cite{tacto}, both the sensor's and object's meshes are primarily introduced in OpenGL \cite{opengl}, subsequently modified within the PyBullet simulation. Within this context, depth maps might be procured from PyBullet or concurrently produced in OpenGL alongside tactile visuals.

As for the physics-based line, \cite{fem1} adopts ABAQUS \cite{abaqus} to simulate the elastomer deformation. However, ABAQUS requires many computational resources (e.g., a cluster) and is hard to integrate into a robot simulator.

\section{IPC Preliminary}
To leverage the advantages of FEM-based modeling and explicit mesh representation, the key challenge here is to make the contact intersection-free and inversion-free. We introduce incremental potential contact (IPC) \cite{ipc} to address it, and will briefly describe its mechanism. %

In a dynamics system, without contact and friction, the dynamics differential equation based on Newton's law for vertices can be formulated as:
\begin{equation}
    \begin{cases}
      f^{ext} + f^{elastic} = M \dot{v}\\
      v = \dot{x}
    \end{cases}
    \label{eqn:dynamics_ode}
\end{equation}
where $f^{ext}$ is the external forces applied to the vertices, and $f^{elastic}$ is the forces due to elastic deformation. $x$ and $v$ is the position and velocity of the vertices, $M$ is the mass matrix. 

Putting Eq. \ref{eqn:dynamics_ode} under the implicit Euler scheme: 
\begin{equation}
    \begin{cases}
      f^{ext}_{t+1} + f^{elastic}_{t+1} = \frac{M}{h}(v_{t+1}-v_{t})\\
      v_{t+1} = \frac{1}{h}(x_{t+1} - x_{t})
    \end{cases}
    \label{eqn:dynamics_euler}
\end{equation}
is equivalent to minimizing the Incremental Potential energy: 
\begin{equation}
    E(x, x_t, v_t) = \frac{1}{2}(x-\hat{x})^TM(x-\hat{x}) + h^2\Phi(x),
    \label{eqn:ip_energy}
\end{equation}
which means $x_{t+1} = arg \min_{x} E(x, x_t, v_t)$. In other words, minimizing the energy can lead to the correct evolution direction in terms of Newton's law (Eq. \ref{eqn:dynamics_ode}). $t$ is the discretized time, $h$ is the time step size, $\Phi(x)$ is the hyper-elastic energy of deformable objects where the relationship $\frac{\partial \Phi}{\partial x} = f^{elastic}$ holds due to the principle of virtual work. $\hat{x}=x_t + hv_t + h^2M^{-1}f^{ext}$. 

Considering contact, to prevent collision, IPC adopts a barrier function:
\begin{equation}
b(d) = b(d, \hat{d})=
\begin{cases}
  -(d - \hat{d})^2log(\frac{d}{\hat{d}}), &0<d<\hat{d}\\
  0 &d\geq \hat{d}
\end{cases}
\end{equation}
where $d$ is the geometric distance between contact primitive pairs including point-triangle pairs and edge-edge pairs, and $\hat{d}$ is a threshold parameter. 

With this barrier function $b(\cdot)$, one can construct a barrier energy term:
\begin{equation}
B(x) = \kappa\sum_{k\in C} b(d_k(x))
\end{equation}
which induces the collision forces between contact primitive pairs, where $C$ represents the set of any primitive pair $k$ (i.e., an edge-edge pair or a point-triangle pair in the 3D case) with its geometric distance $d_k$ less than $\hat{d}$, $\kappa$ is the weight parameter for this barrier energy term.

Considering the friction, IPC uses a $C^1$ smooth approximation of the Coulomb friction model
\begin{equation}
f_k = -\mu\lambda_k f_1(\lVert u_k\rVert) T_k(x)\frac{u_k}{\lVert u_k\rVert}
\end{equation}
where $\mu$ represents the coefficient of friction, $\lambda_k=\kappa\lVert\frac{\partial b(d_k(x))}{\partial x}\rVert$ is the contact force magnitude, $T_k(x)$ represents the sliding basis, $u_k=T_k(x)^T(x-x_t)$ is the tangential relative displacement vector at the local orthogonal frame. Here, $f_1(x)$ is a $C^1$ smooth function
\begin{equation}
f_1(x)=
\begin{cases}
-\frac{y^2}{\epsilon_v^2h^2} + \frac{2y}{\epsilon_vh}, &y\in(0, h\epsilon_v)\\
1, & y\geq h\epsilon_v
\end{cases}
\end{equation} making the friction force $f_k$ integrable, characterizing the transition between dynamic friction and static friction. Now if we lag the sliding basis and contact force magnitude to values $T_k^n$, $\lambda_k^n$ solved in the last optimization step, then we can derive the variational frictional energy by integrating $f_k$: 
\begin{equation}
D_k(x, x_t) = \mu\lambda_k^nf_0(\lVert u_k\rVert)
\end{equation} which satisfies $f_k(x)=-\nabla_xD_k$, where $f_0(x)=\int_{\epsilon_vh}^x f_1(s) ds + \epsilon_vh$ and $u_k=T_k^n(x)^T(x-x_t)$. Here $\epsilon_v > 0$ is a threshold parameter to handle the transition between dynamic and static friction.  
So the total lagged variational frictional energy would be $D(x, x_t) = \sum_{k\in C} D_k(x, x_t)$, and we finally solve frictional contact by minimizing the energy and updating lagged sliding basis along with contact force magnitude alternately. 

After adding the friction energy $D(x, x_t)$ and the barrier energy $B(x)$ above into the incremental potential energy $E(x, x_t, v_t)$, one can solve a dynamics system with contact and friction by minimizing the total energy $\hat{E}=E(x, x_t, v_t) + B(x) + D(x, x_t)$ using the Projective Newton optimization algorithm. During each Newton step, to ensure that no penetration happens, IPC applies the Continuous Collision Detection (CCD) algorithm to compute the Time Of Impact (TOI) and subsequently uses it to filter the Newton step size. In this way, IPC could guarantee that its simulation results are intersection-free and inversion-free. 

\section{TacIPC}\label{sec:tacipc}
\begin{figure}[t!]
\vspace{0.2cm}
    \centering
    \includegraphics[width=1\linewidth]{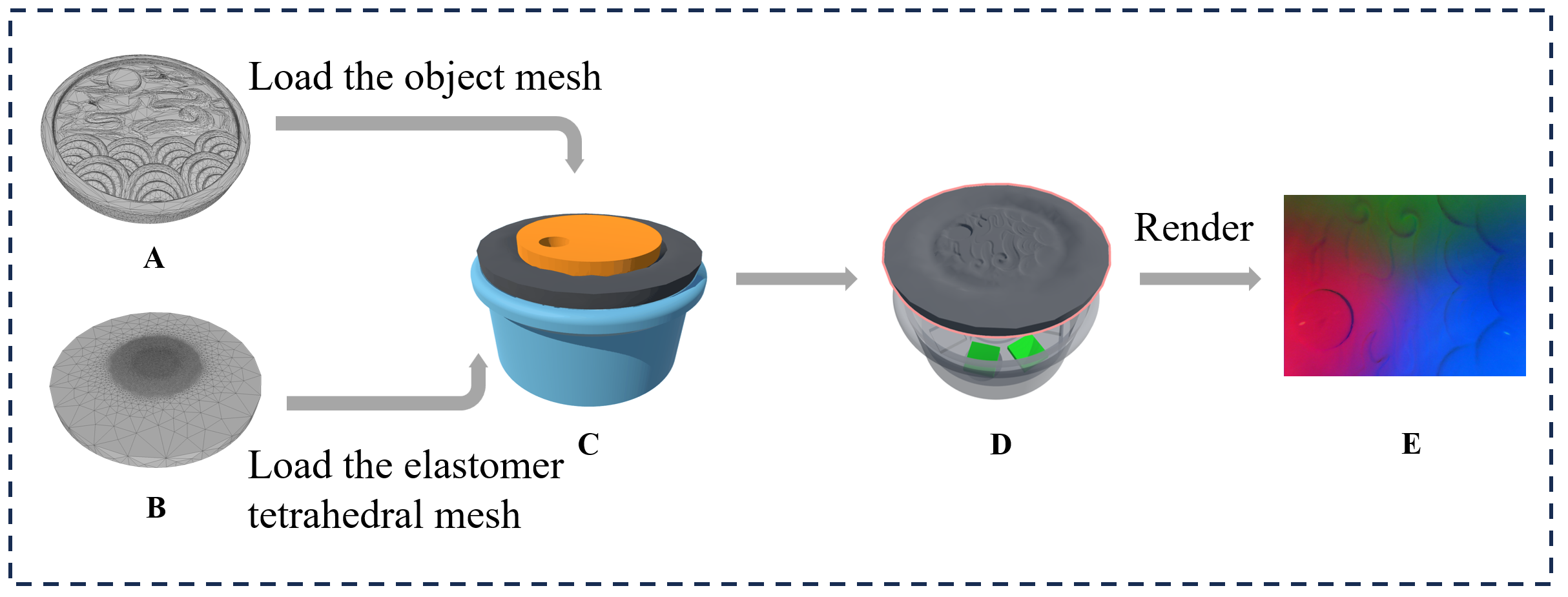}
    \caption{TacIPC: Simulation and Rendering Pipeline. The object mesh (A) and the elastomer tetrahedral mesh (B) are loaded into the IPC simulation scene (C). The elastomer deformation solved by IPC is used to render the pseudo tactile image (E) by an MC-Tac model (D).}
    \label{fig:pipeline}
\end{figure}

By incorporating IPC into tactile simulation, we construct a robot interface for tactile sensing and rendering.
The overall pipeline is illustrated in Fig. \ref{fig:pipeline}.

\subsection{Elastomer Simulation}
In TacIPC, we apply FEM for elastomer simulation. In our work, we use tetrahedral meshes to represent objects. For FEM, previous work \cite{tacchi} has mentioned concerns about the mesh distortions, which may cause irregular meshes or negative element volume and therefore lead to a low accuracy of stress. IPC partially solves this problem by applying CCD and optimization step size filtering techniques to eliminate negative volume and mesh intersection issues, resulting in a higher simulation quality. Additionally, the typical use cases of tactile sensors will not cause large elastomer deformation, thus these problems are usually not significant.

For elastomer simulation, we model the elastomer with a tetrahedral mesh with $K$ vertices and $N$ tetrahedra. When the elastomer is pressed by an external surface which is also represented by mesh, we apply IPC to handle the contact. The numbers of $K$ and $N$ depend on the mesh discretization strategy of the gel elastomer. We discuss different strategies in Sec. \ref{sec:ablation}. In our work, the main experiments are conducted with strategy: $K=23661$ and $N=104675$, in the central region of the mesh where most of the contact happens, vertices and tetrahedral cells reach the highest density where the average edge length is 0.1mm.

\subsection{Connection With A Robot Simulator}

In Tacchi \cite{tacchi}, it shares only the contact object velocity with the external simulator such as MuJoCo \cite{mujoco}. Due to the different contact modeling in Tacchi (particle for object representation, MPM for physics simulation) and MuJoCo (mesh for object representation, penalty-based optimization \cite{mujoco_opt} for physics simulation), the tactile simulation and robot simulation are independent. That is, the contact force between the object and tactile sensor in these simulators will be different, and the particle locations of the gel elastomer in Tacchi will not align with the vertex locations in the external simulator.

However, in TacIPC, we can simply address such issues by sharing all vertex velocities and locations with an external simulator supporting mesh modeling such as RFUniverse.

\subsection{Rendering}
Since rendering is related to the lighting design of the optical tactile sensors, different sensors may have different designs. We adopt the MC-Tac sensor \cite{mctac} for the main experiments, as it is an open-source optical tactile sensor.

We load an MC-Tac sensor model in Unity where light source positions and parameters (e.g., lighting color temperature) are aligned with those of the real-world sensor. We also align the Unity camera pose and parameters with those of the real-world camera in the sensor. The deformed elastomer surface generated by TacIPC reflects the light emitted by these light sources, so tactile images can be captured by the Unity camera and subsequently rendered by the standard Unity physics-based rendering pipeline (e.g., real-time ray-tracing). Finally, we subtract the simulation reference image from the rendered image and then add it to the reference image collected from the real world to obtain the result. It can introduce real-world sensory noise and further reduce the sim-to-real gap.

\section{Experimental setup}
In this section, we introduce the task settings to validate the ability of TacIPC in Sec. \ref{sec:task}, real-world setup in Sec. \ref{sec:real_setup}, and simulation setup in Sec. \ref{sec:virtual_setup} respectively. 

\subsection{Benchmark Tasks}\label{sec:task}
We collect both real-world data and simulation data to evaluate TacIPC in 3 tasks: (1) pseudo-image quality assessment. In this task, we compare the tactile images generated by the simulator with ground truth collected from the real world under the Structural Similarity (SSIM), Mean Absolute Error (MAE), and Peak Signal-to-Noise Ratio (PSNR) metrics. TacIPC and other simulator baselines including the depth-based method \cite{depth1} and Tacchi \cite{tacchi} are evaluated respectively. (2) marker displacement prediction. In this task, we first press the contact object onto the tactile sensor elastomer, then use a robot gripper to rotate and push the contact object respectively. During the process, we record the movement of the markers on the elastomer. We align all the movements of the contact object in Tacchi and TacIPC respectively with those in the real-world experiments. Finally, we compare the marker displacement predicted by Tacchi and TacIPC with the ground truth collected from real-world experiments. (3) deformed geometry estimation. In this task, we train a U-Net \cite{unet} to predict depth images of contact objects from the corresponding tactile images. Two networks are trained on synthetic datasets generated by Tacchi \cite{tacchi} and TacIPC and tested on real-world data. We assess the sim-to-real gap of their output through MAE and Mean Squared Error (MSE).

\begin{figure}[h!]
    \centering
    \includegraphics[width=1\linewidth]{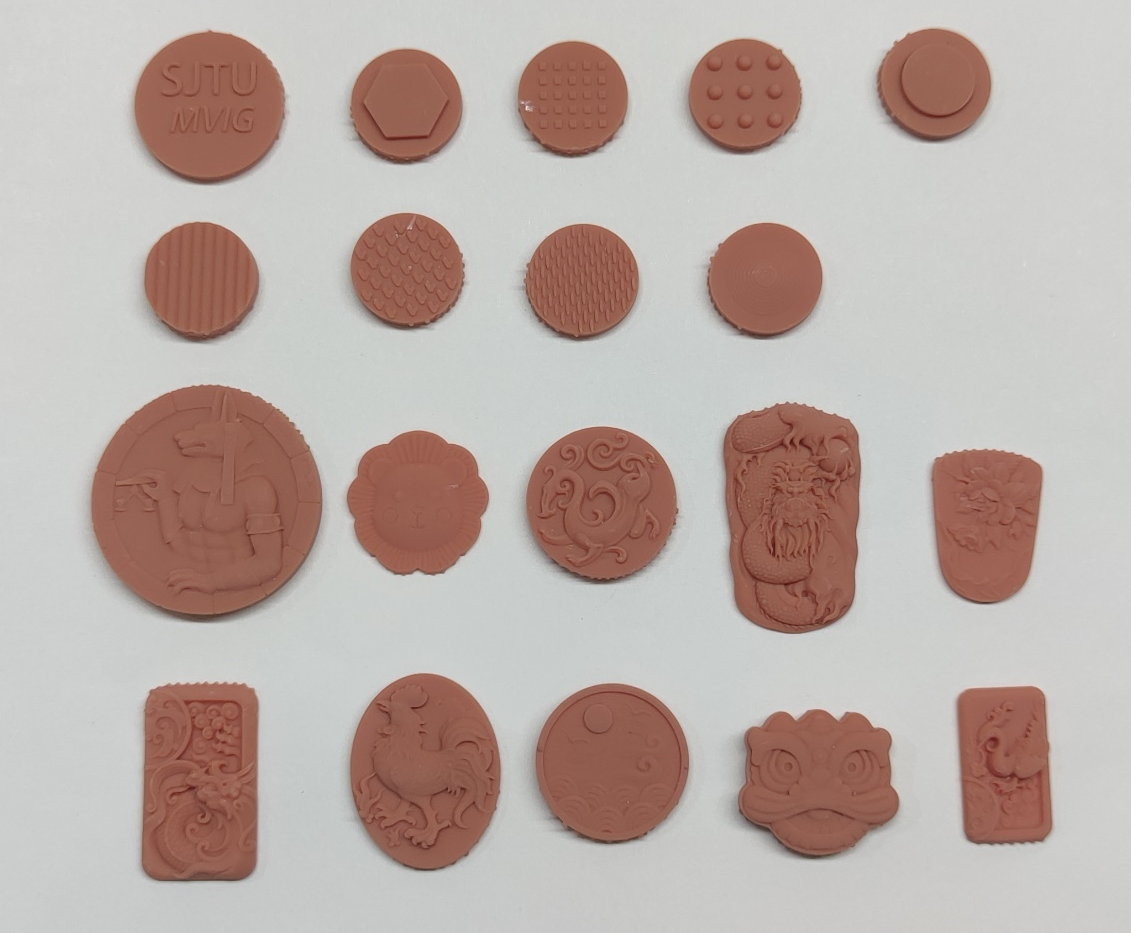}
    \caption{3D printed objects with various complex patterns.}
    \label{fig:printed_indenters}
\end{figure}

\begin{figure*}[t!]
    \centering

    \includegraphics[width=1\linewidth]{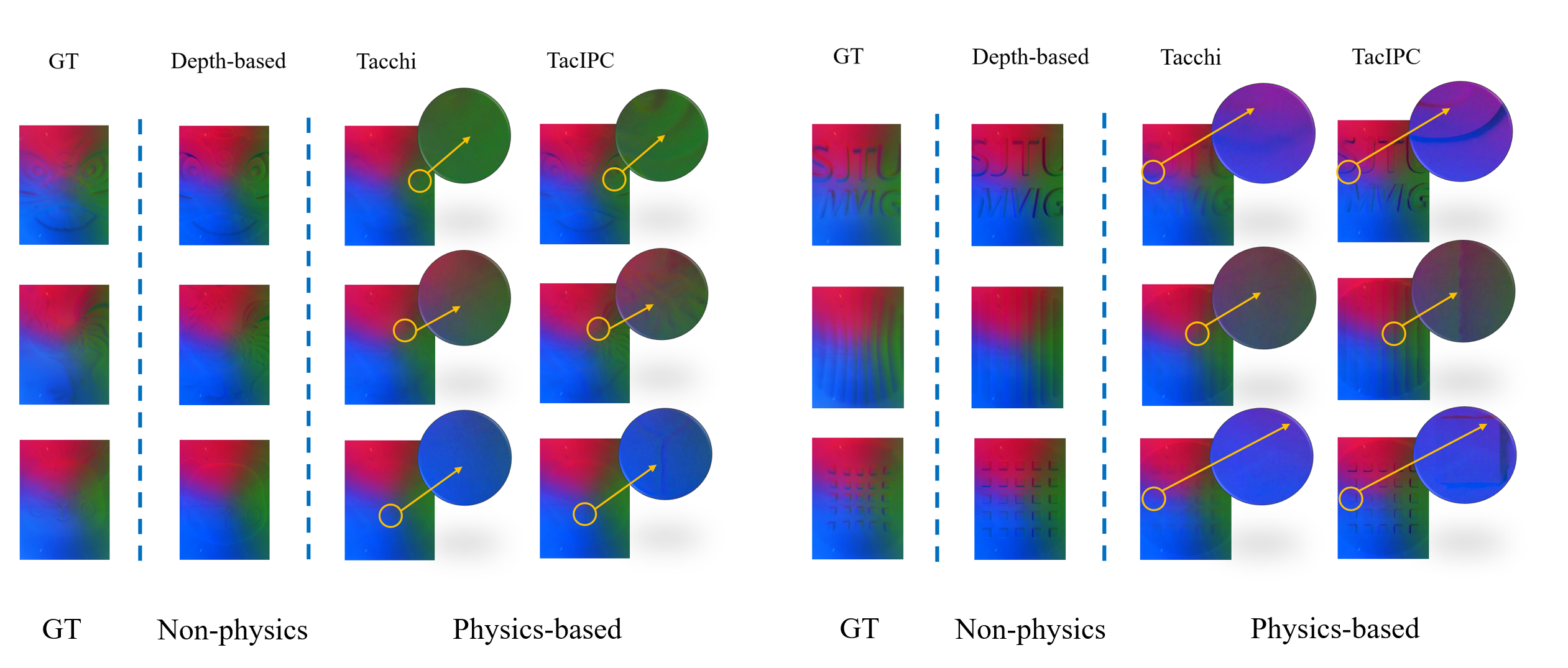}

    \caption{Pseudo images of the depth-based method\cite{depth1} (2nd column), Tacchi\cite{tacchi} (3rd column), and TacIPC (4th column). Ground truth (1st column) is collected from real-world experiments. Within physics-based methods, TacIPC generates results with more detailed texture and higher quality than those of Tacchi. }
    \label{fig:pseudo_baseline}
\end{figure*}

\subsection{Real-world Setup}\label{sec:real_setup}
We obtained an object dataset by 3D printing 19 object meshes where 9 of them have different regular patterns such as alphabet letters and polygons and the rest of them have various detailed textures. As from Fig. \ref{fig:printed_indenters}, all of these objects are short cylinders similar to coins, where the patterns are printed on one side of them. We glued the other side of these objects to a wooden cube when conducting the experiments. A Flexiv robot with an AG-95 gripper grasps the cube, and then it will move to the MC-Tac sensor which is fixed on a horizontal tabletop. Subsequently, the gripper goes down to press the elastomer of the sensor. Meanwhile, the tactile snapshots are captured by the MC-Tac camera.

\subsection{Virtual-world Setup}\label{sec:virtual_setup}
In the TacIPC simulator, we first use ABAQUS \cite{abaqus} to apply tetrahedral meshing on the MC-Tac elastomer triangle mesh which is a cylinder with a radius of 15mm and a thickness of 2mm. The average edge length of the contact area is about 0.1mm. The vertex number and cell number of the tetrahedral mesh are 23661 and 104675 respectively. For the object mesh, we apply the quadric edge collapse decimation mesh simplification algorithm \cite{simplification} provided by the MeshLab software \cite{meshlab} before sending them into the simulation scene, resulting in meshes each with $\sim$8000 vertices, since most of the original meshes have more than 500,000 vertices which are too many for the simulation and will dramatically increase the computation cost with no apparent quality improvement. In our TacIPC simulation, the timestep $dt=0.01s$, distance threshold for barrier energy $\hat{d}$ is set to be $1\times 10^{-3}$ times of the diagonal length of the simulation scene, barrier energy weight $\kappa=10^6$. For the contact object, we model it as a rigid body with density $\rho=1\times10^3kg/m^3$. As for the MC-Tac gel elastomer, we measure the gel material parameters in the real world and set them in the simulation scene. These parameters are as follows: gel density $\rho_{gel}=1.01\times10^3kg/m^3$, Young's modulus $E_{gel}=1.23 \times 10^5 Pa$, and Poisson's ratio $\nu_{gel}=0.43$. In the simulation, we use the augmented Lagrangian algorithm to fix the side of the elastomer which is glued to a horizontal surface in the real world and to control the movement of the contact object to press the elastomer. In this way, we can save the deformation results of the elastomer and subsequently use them to render tactile images in Unity.

\begin{figure}[ht!]
    \centering

    \includegraphics[width=1\linewidth]{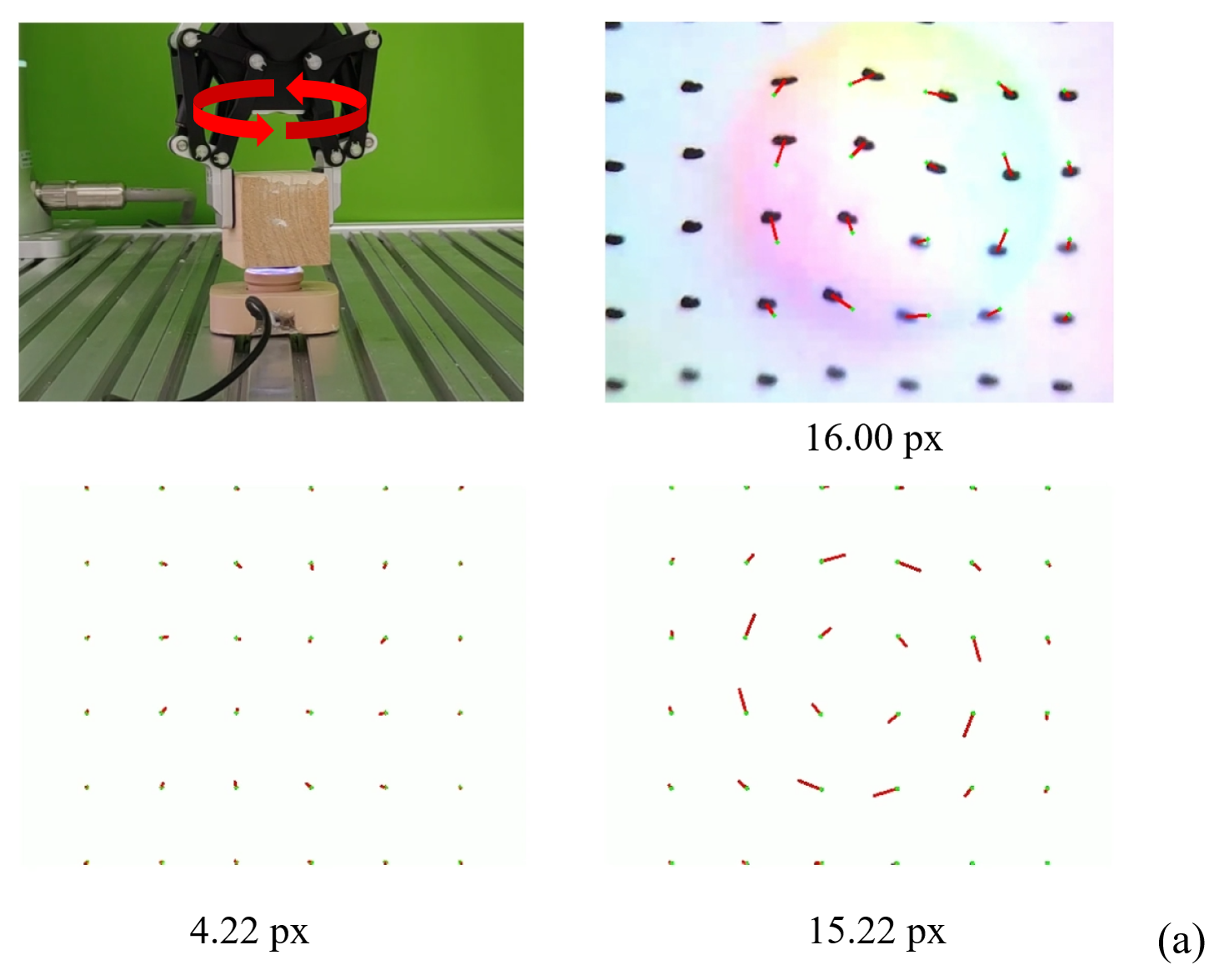}
    \includegraphics[width=1\linewidth]{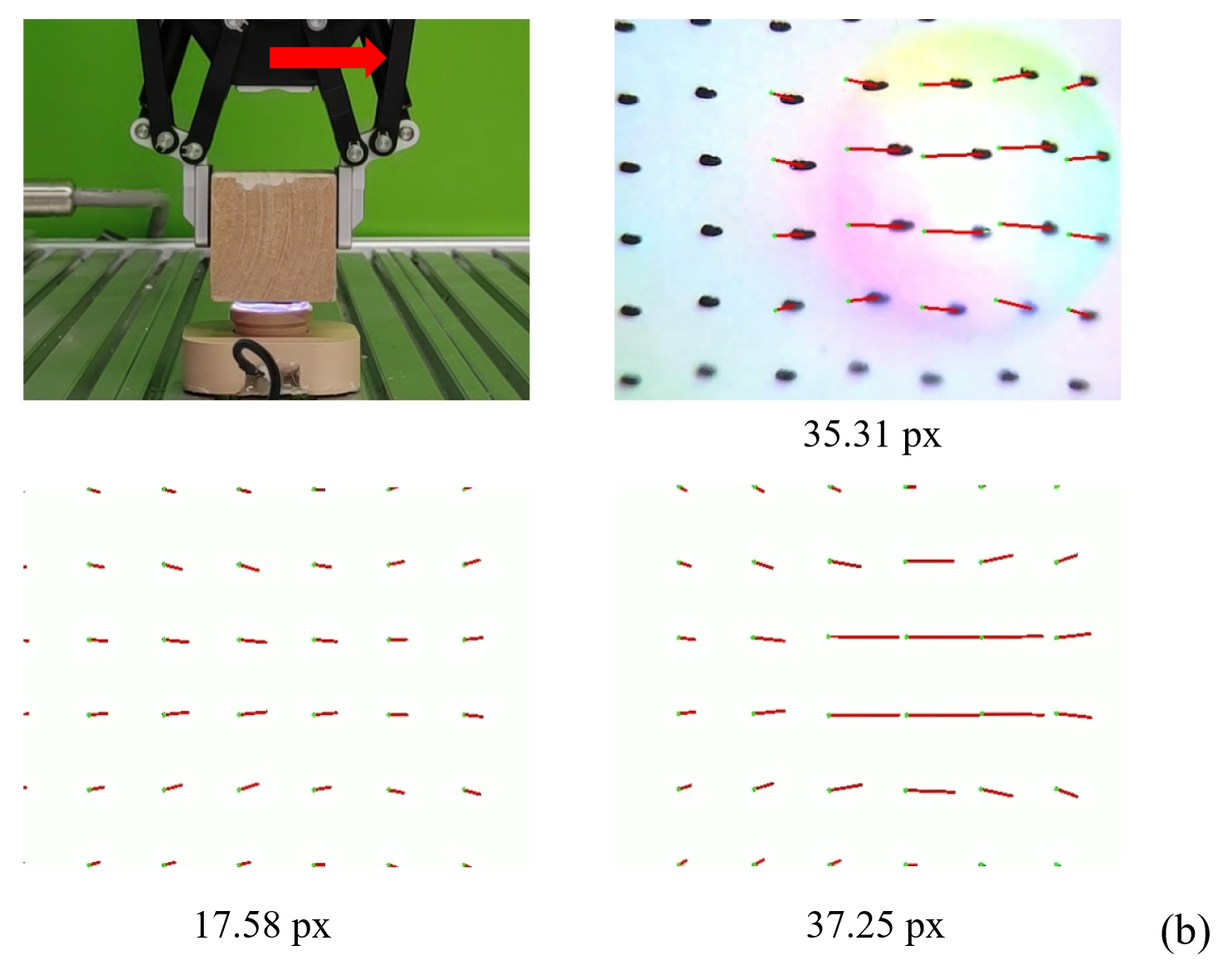}
        \caption{Marker displacement in rotational (a) and shearing (b) friction experiments. In both (a) and (b), the top left image illustrates the gripper motion during the real-world experiments. The top right image shows the marker displacement recorded in the real world. The bottom left image and the bottom right image show the marker displacement predicted by the Tacchi and the TacIPC simulator respectively. The red line represents the displacement vectors of markers, where the end with a green dot on each line represents the initial marker position, and the position of the other end is the final marker position. Pixel values under the images are the average displacement of 20 selected markers which correspond to the 20 green dots in the top right sub-images of (a) and (b).}
        \label{fig:marker_disp}
\end{figure}
\section{Experiments}

\subsection{Pseudo-Image Quality Assessment}
To evaluate the pseudo image quality of our method, we collect tactile sensor images of 19 3D-printed objects with complex textures in the real world. The printed objects and their meshes are shown in Fig. \ref{fig:pseudo_baseline}. %
As Fig. \ref{fig:pseudo_baseline} shows, Tacchi generates over-smooth results due to its depth field interpolation step. While pseudo images generated by TacIPC illustrate detailed and clear complex textures of contact objects. However, such over-smooth effects may be not properly reflected in the pixel-wise metrics like MAE, and PSNR reported in TABLE \ref{tab:sim_tac_img}. It inspires us to adopt metrics on physics, as described in the next section.

\begin{table}[ht!]
    \centering
    \begin{tabular}{c|cccc|c}
    \hline
       Simulation  & SSIM$\uparrow$ & MAE$\downarrow$ & PSNR$\uparrow$    \\\hline
       depth-based  & 0.86 & 0.043 & 55.92   \\ \hline
       Tacchi & \textbf{0.90} & \textbf{0.034} & \textbf{59.17}  \\ \hline
       TacIPC & \textbf{0.90} & 0.037 & 57.72  \\\hline
    \end{tabular}
    \caption{Pseudo image quality comparison. Tacchi and TacIPC generates better results in terms of SSIM. In terms of MAE and PSNR, Tacchi achieves the best results among the three simulation methods, but generating over-smooth results. }
    \label{tab:sim_tac_img}
\end{table}

\subsection{Marker Displacement Prediction}
 As \cite{gelsight} shows, the marker displacement of tactile sensors is crucial since it could be used to predict force distribution, contributing to more stable manipulation policies. Here we conduct experiments to compare the marker displacement computed by different physics simulation methods and that in the real world. To fully test the marker displacement accuracy in different cases, we conducted the experiment under both shearing frictional contacts and rotational frictional contacts. In both settings, the elastomer was pressed to a depth of 0.5mm by the contact object. Subsequently, in the former setting, the gripper will move straight horizontally, causing the gel elastomer to shear, while in the latter setting the gripper will rotate around a vertical axis through its center, causing the gel elastomer surface to rotate. %

\begin{figure}
    \centering

    \includegraphics[width=0.8\linewidth]{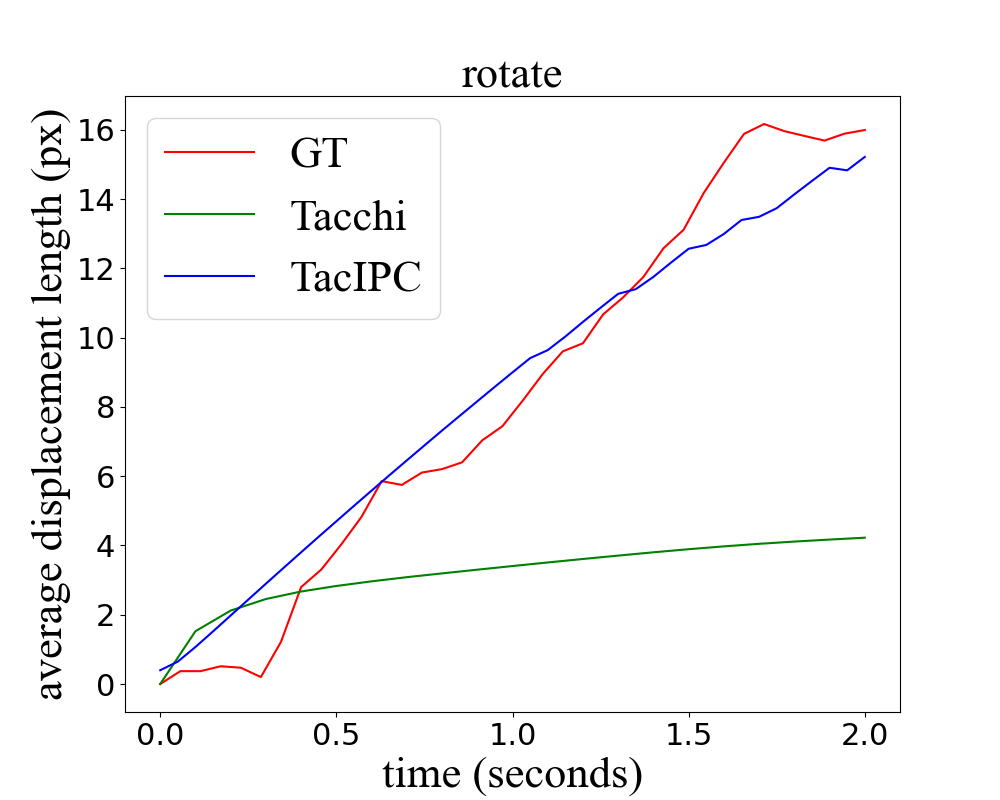}
    \includegraphics[width=0.8\linewidth]{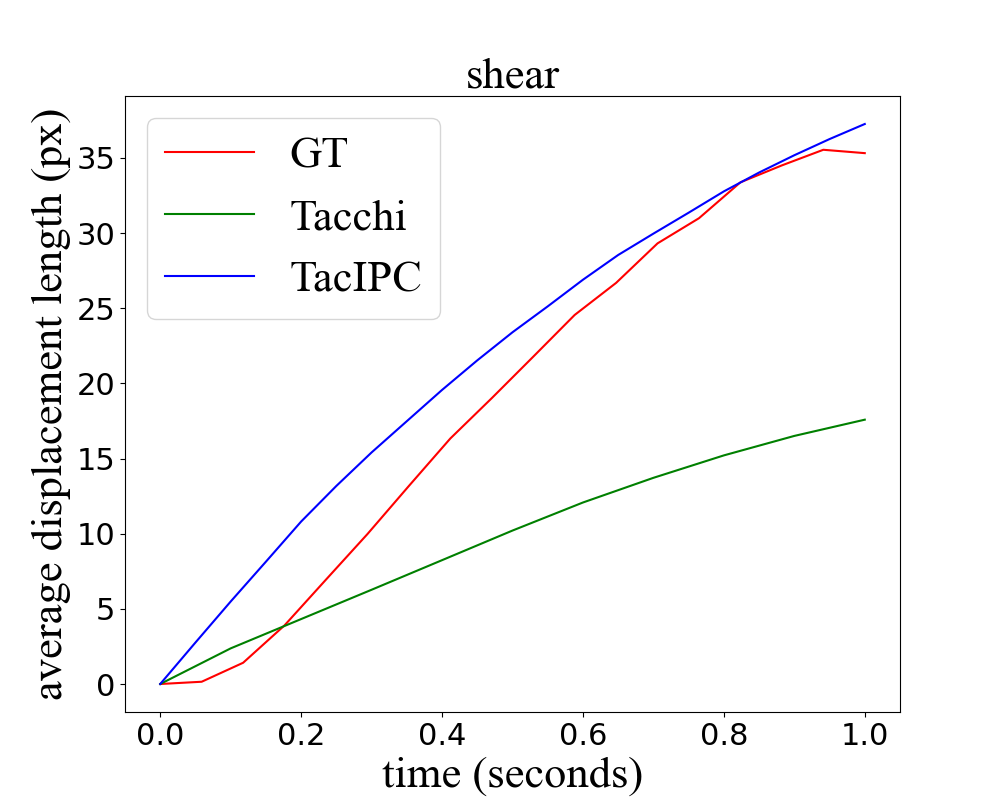}

    \caption{Curves of average marker displacement length in the real world, Tacchi simulator, and TacIPC simulator respectively. Results generated by TacIPC show higher consistency with the real-world data than those of Tacchi. }
    \label{fig:marker_disp_curves}
\end{figure}

We record the marker displacement in both settings in the real world, Tacchi simulator, and TacIPC simulator respectively, as shown in Fig. \ref{fig:marker_disp}. Take the displacement in the real world as the ground truth, marker displacement in the Tacchi simulator is much shorter than the ground truth, while marker displacement in TacIPC is closer to those in the real world. We also plot the accumulated average displacement length of 20 markers in the center of the contact region in the real world, Tacchi simulator, and TacIPC simulator for each frame in Fig. \ref{fig:marker_disp_curves}. It clearly illustrates that the marker displacement in our TacIPC is much more consistent with the real world in a long-term sequence than the MPM-based Tacchi simulator.

\subsection{Deformed Geometry Estimation} \label{sec:deformed_geom_est}
\begin{figure}
    \centering

    \includegraphics[width=1\linewidth]{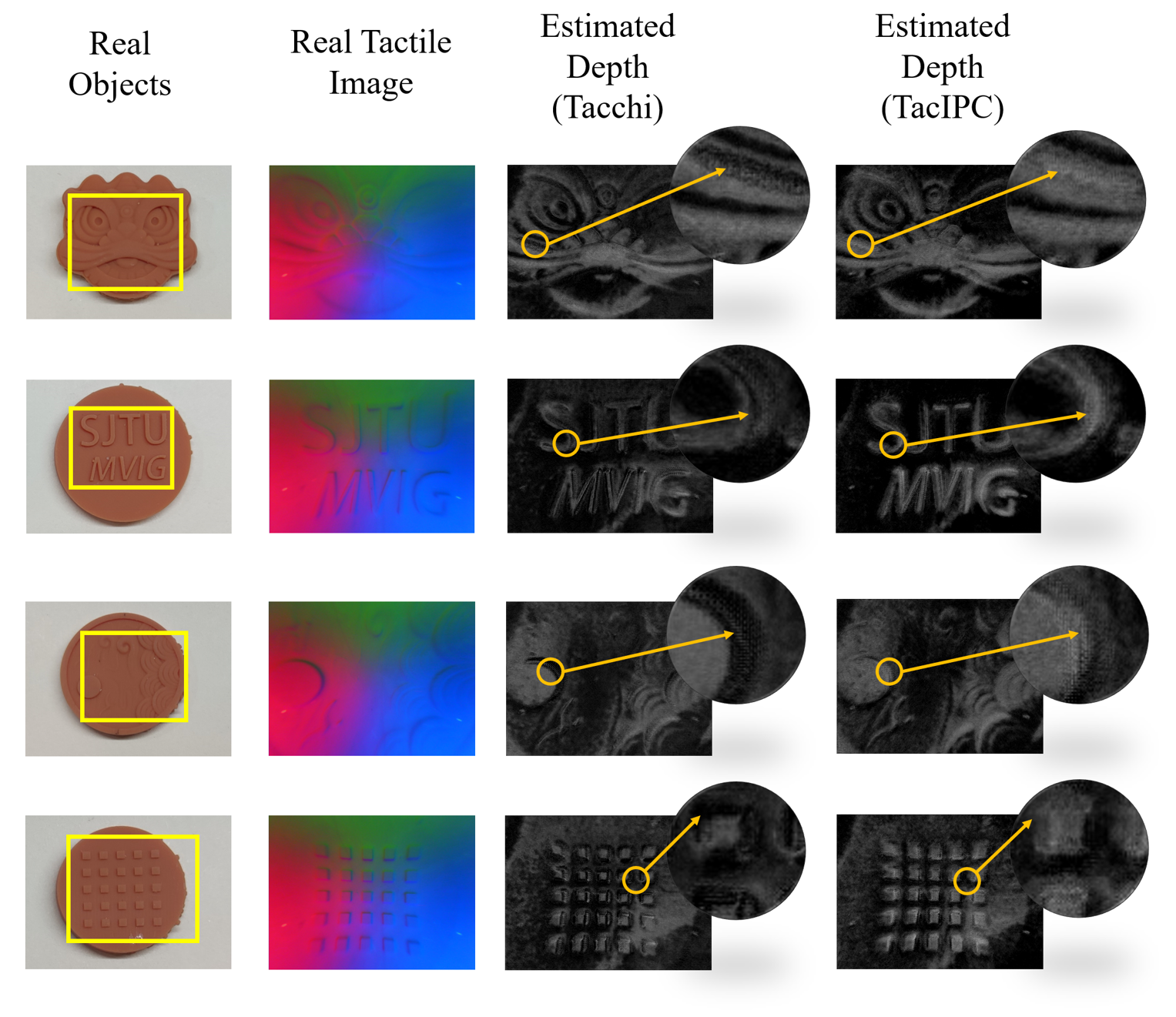}

    \caption{Validation results of the depth estimation models trained by synthetic datasets generated by Tacchi (3rd column) and TacIPC (4th column) respectively. From the zoomed-in details, we can observe that the Tacchi depth estimation model tends to predict large depth values for the shadow part caused by object edges. While the TacIPC model does not share this artifact and generates more realistic results.}
    \label{fig:depth_estimation}
\end{figure}

We train a U-Net to estimate the depth map of the contact object given its tactile image. We collect 20 tactile images for each of the 24 object meshes with various random poses to obtain the training dataset by using our TacIPC simulator. We first train the network on this dataset and then validate it on real images collected from the real world to examine the sim-to-real gap between the TacIPC simulator and the real world, which is reported in Table \ref{tab:depth_est_real}. The reference depth in the real world is obtained by aligning the flat part of the reconstructed object surface with that of the ground truth contact object mesh in Unity. Qualitative results are shown in Fig \ref{fig:depth_estimation}. The high quality of the estimated depth images illustrates the gap between simulation and real-world is partially bridged by the accuracy of TacIPC. 

\begin{table}[ht!]
\vspace{0.4cm}
    \centering
    \begin{tabular}{c|cccc|c}
    \hline
       Simulation  & MAE$\downarrow$ & MSE$\downarrow$  \\\hline
       Tacchi & 0.03072 &  0.001423  \\ \hline
       TacIPC &  \textbf{0.02708} & \textbf{0.001254}  \\\hline
    \end{tabular}
    \caption{Depth image reconstruction error on real-world data of the two U-Nets trained on Tacchi and TacIPC respectively. U-Net trained on the TacIPC-generated dataset achieves higher accuracy in terms of MAE and MSE.}
    \label{tab:depth_est_real}
\end{table}

\subsection{Ablation Study}\label{sec:ablation}

\begin{figure}
\vspace{0.2cm}
    \centering

    \includegraphics[width=1\linewidth]{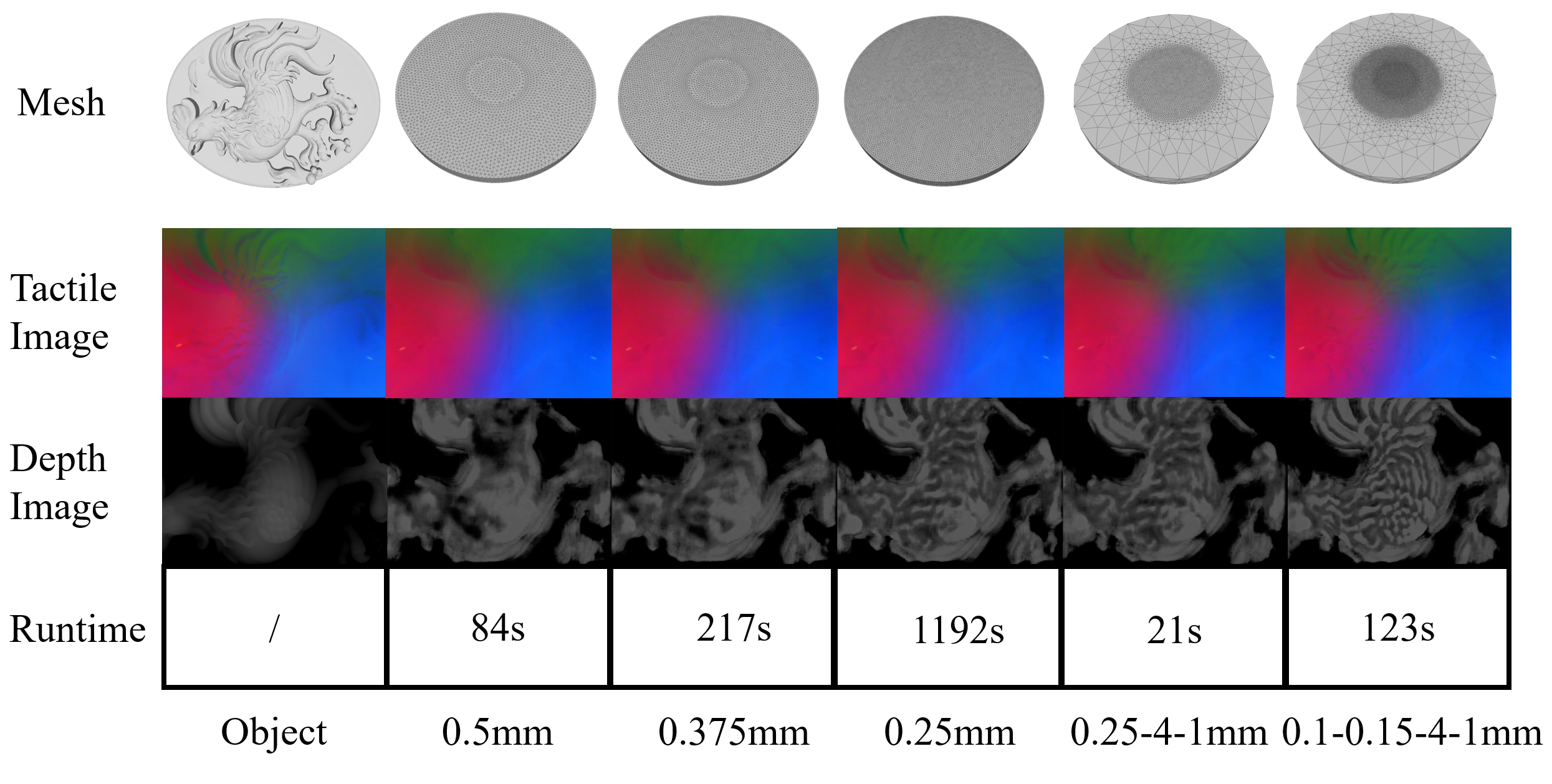}

    \caption{Simulation tactile image and the corresponding predicted depth images comparison among different mesh discretizations. The fourth row illustrates the runtime of tactile image generation using different sensor elastomer mesh discretizations. }
    \label{fig:adaptive_meshing_results}
\end{figure}

\paragraph{Different Meshing Methods}
We test different tactile sensor elastomer tetrahedral mesh discretizations generated by uniform meshing and adaptive meshing techniques by using them to generate tactile images and estimate contact object depth maps using these generated images. Three uniformly discretized meshes and two adaptively discretized meshes are listed in the first row of Fig. \ref{fig:adaptive_meshing_results}. By uniform meshing, we mean discretizing the object with almost uniform edge lengths. In the 2nd column to the 4th column of Fig. \ref{fig:adaptive_meshing_results}, we use average edge lengths of 0.5mm, 0.375mm, and 0.25mm respectively. By adaptive meshing, we mean the density of vertices reaches the maximum, in other words, the average edge length reaches the minimum, around the contact-rich region which is the central region of the elastomer front surface. For the mesh listed in the 5th column of Fig. \ref{fig:adaptive_meshing_results}, the average edge length of the front central region, of the front edge region, and of the back side, is 0.25mm, 4mm, and 1mm respectively. The adaptive mesh placed in the 6th column of Fig. \ref{fig:adaptive_meshing_results} has an average edge length that gradually increases from 0.1mm to 0.15mm and then to 4mm, as the region moves from the front central part to the front edge part. The back side of the mesh has an edge length of 1mm. Fig. \ref{fig:adaptive_meshing_results} also shows the tactile images generated by these discretized meshes and the corresponding depth maps estimated by the U-Net previously described in Sec. \ref{sec:deformed_geom_est}.  From the results we observe that to achieve similar tactile image quality, adaptive meshing needs far fewer vertices and cells than uniform meshing. In all other experiments, we use the adaptive mesh discretization illustrated in the most right column of Fig. \ref{fig:adaptive_meshing_results} for the MC-Tac elastomer.

\paragraph{Computational Cost}
We must admit that in exchange for higher accuracy taking the frictional contacts into account, the computation cost for IPC to simulate the tactile sensor elastomer deformation is much higher than that of explicit MPM. Typically TacIPC needs $\sim$6GB GPU memory to compute the elastomer deformation during 5 large timesteps ($h=0.01s$) when the elastomer is pressed by a simplified object mesh with $\sim$8000 vertices and $\sim$16000 edges. Here the elastomer mesh discretization has 23661 vertices and 104675 tetrahedral cells. All the simulation experiments were running on a GeForce RTX 4090 graphics card. 

\section{Conclusion And Discussion}
We propose TacIPC, an intersection- and inversion-free FEM-based elastomer simulation for optical tactile sensors. TacIPC simulates the deformation of the gel elastomer to generate high-quality tactile images. It can also accurately predict the marker displacement of tactile sensors by applying IPC to properly handle contact and friction. Additionally, we train a depth estimation model that is able to reconstruct the contact object geometry on a TacIPC-generated synthetic dataset, showing a reduced sim-to-real gap. Moreover, TacIPC can be integrated with existing simulators supporting mesh modeling. 

In this work, an efficient and standard rendering model is used currently. To gain more realistic tactile images, one needs to improve the rendering model and apply techniques to calibrate its parameters. We leave these for future work.  

\bibliographystyle{IEEEtran}
\bibliography{reference}

\end{document}